\documentclass{article}

\usepackage[preprint]{neurips_2025}

\usepackage[utf8]{inputenc} 
\usepackage[T1]{fontenc}    
\usepackage{hyperref}       
\usepackage{url}            
\usepackage{booktabs}       
\usepackage{amsfonts}       
\usepackage{tcolorbox}      
\usepackage{nicefrac}       
\usepackage{microtype}      
\usepackage{xcolor}         

\hypersetup{
    colorlinks=true,
    linkcolor=blue,
    filecolor=blue,      
    urlcolor=blue,
    citecolor=cyan,
}
\usepackage{colortbl}
\usepackage{color}
\usepackage{cite}
\usepackage{times}
\usepackage{latexsym}
\usepackage{array}
\usepackage{graphicx}
\usepackage{epstopdf}
\usepackage{makecell}
\newcolumntype{L}[1]{>{\raggedright\let\newline\\\arraybackslash\hspace{0pt}}m{#1}}
\newcolumntype{C}[1]{>{\centering\let\newline\\\arraybackslash\hspace{0pt}}m{#1}}
\newcolumntype{R}[1]{>{\raggedleft\let\newline\\\arraybackslash\hspace{0pt}}m{#1}}
\usepackage{multirow}
\usepackage{bbding}
\usepackage{pifont}
\usepackage{amsmath}
\usepackage{arydshln}
\usepackage{cleveref}
\usepackage{wrapfig}
\usepackage{enumitem}
\crefname{section}{§}{§§}
\Crefname{section}{§}{§§}
\usepackage{bm}

\newcommand{\Xmat}[0]{{{\bf X}}}

\definecolor{darkgreen}{rgb}{0, 0.5, 0.0}

\newcommand{\token}[0]{\textcolor{darkgreen}{[X] }}
\newcommand{\doctoken}[0]{\textcolor{orange}{[D] }}

\title{Beyond Hard and Soft: Hybrid Context Compression for Balancing Local and Global Information Retention}

\author{%
    \textbf{Huanxuan Liao}$^{\spadesuit\clubsuit}$,
    \textbf{Wen Hu}$^{\diamondsuit}$,
    \textbf{Yao Xu}$^{\spadesuit\clubsuit}$,
    \textbf{Shizhu He}$^{\spadesuit\clubsuit}$\thanks{Corresponding author},
    \textbf{Jun Zhao}$^{\spadesuit\clubsuit}$,
    \textbf{Kang Liu}$^{\spadesuit\clubsuit}$
    \\
    $^{\spadesuit}$Institute of Automation, Chinese Academy of Sciences \\
    $^{\clubsuit}$University of Chinese Academy of Sciences ~~~
    $^{\diamondsuit}$Ant Group \\
    {\texttt{liaohuanxuan2023@ia.ac.cn}} \\%
}

\begin{document}

\maketitle

\begin{abstract}

Large Language Models (LLMs) encounter significant challenges in long-sequence inference due to computational inefficiency and redundant processing, driving interest in context compression techniques. Existing methods often rely on token importance to perform hard local compression or encode context into latent representations for soft global compression. However, the uneven distribution of textual content relevance and the diversity of demands for user instructions mean these approaches frequently lead to the loss of potentially valuable information.
To address this, we propose \textbf{Hy}brid \textbf{Co}ntext \textbf{Co}mpression (HyCo$_2$) for LLMs, 
which integrates both global and local perspectives to guide context compression while retaining both the essential semantics and critical details for task completion.
Specifically, we employ a hybrid adapter to refine global semantics with the global view, based on the observation that different adapters excel at different tasks. 
Then we incorporate a classification layer that assigns a retention probability to each context token based on the local view, determining whether it should be retained or discarded. 
To foster a balanced integration of global and local compression, we introduce auxiliary paraphrasing and completion pretraining before instruction tuning. This promotes a synergistic integration that emphasizes instruction-relevant information while preserving essential local details, ultimately balancing local and global information retention in context compression.
Experiments show that our HyCo$_2$ method significantly enhances long-text reasoning while reducing token usage. It improves the performance of various LLM series by an average of 13.1\% across seven knowledge-intensive QA benchmarks. Moreover, HyCo$_2$ matches the performance of uncompressed methods while reducing token consumption by 88.8\%. 
Our code will be available at \url{https://github.com/Xnhyacinth/HyCo2}.


\end{abstract}

\vspace{-0.4cm}
\section{Introduction}

Large Language Models (LLMs) \citep{gpt4, LLaMA3, qwen25} demonstrate strong performance across diverse real-world tasks, particularly those requiring the processing of extensive text inputs \citep{liu2025comprehensive}, such as documents and literature \citep{silver, 128k, extending}.
Handling extended context is essential for advanced applications like retrieval-augmented generation (RAG) \citep{hipporag, longrag}, long-term memory systems \citep{hiagent, memoryllm}, and complex reasoning frameworks \citep{neural, let}.
However, supporting such capabilities often requires processing prompts (including instruction, documents, examples, thought process, etc.) containing tens of thousands of tokens \citep{awakening}, which presents significant challenges. Primarily, the quadratic complexity of the attention mechanism \citep{xia-etal-2024-unlocking} leads to escalating computational and financial costs. It also weakens the model's capacity to provide relevant information when addressing specific tasks,
particularly in the presence of noisy or overly lengthy inputs \citep{lost, shi2023large, InstructingLL}. Furthermore, LLM architectures typically enforce strict context window limitations, imposing explicit upper bounds on input size.

\begin{figure}[t]
\centerline{\includegraphics[width=1.0\textwidth]{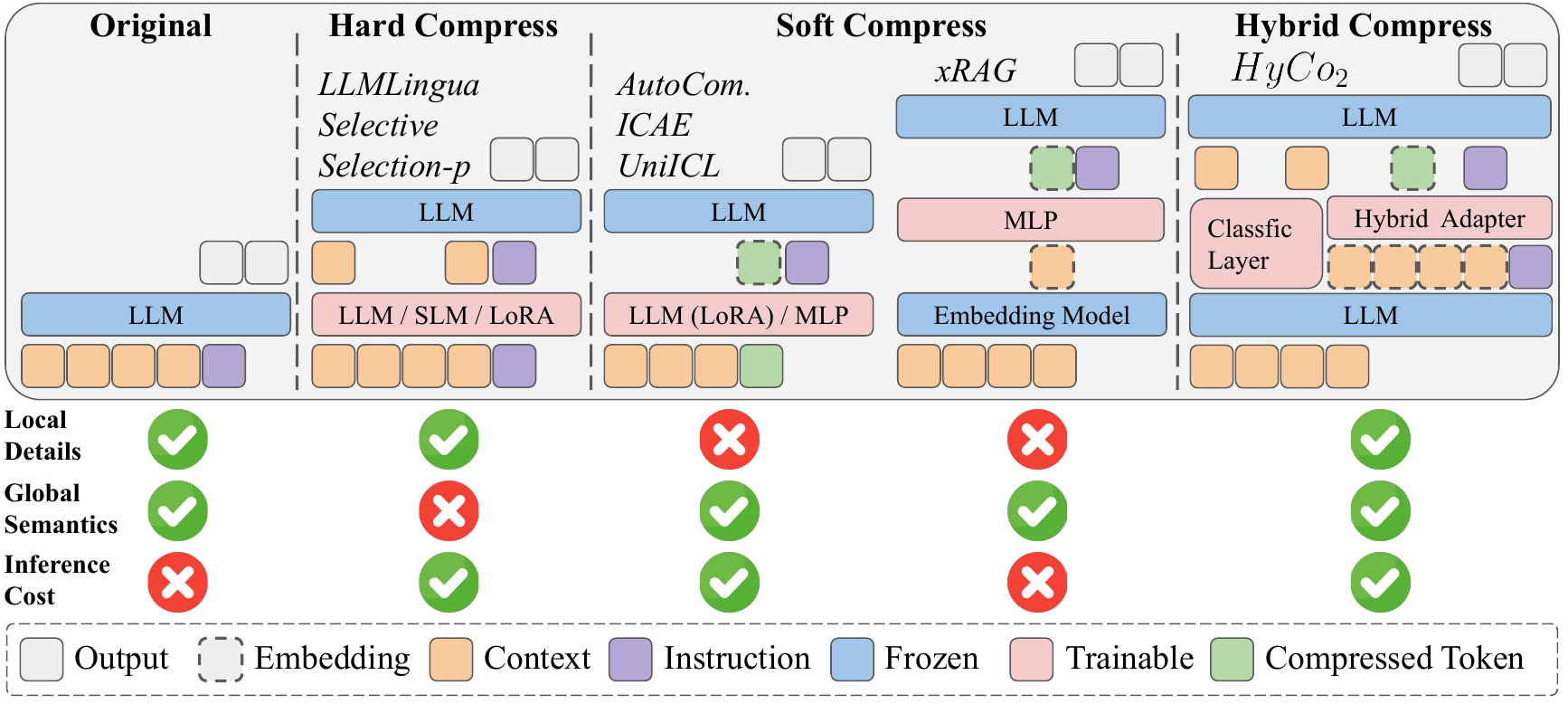}}
\caption{Different paradigms for processing long-text inputs: (a) original input, (b) hard compression, (c) soft compression and (d) our hybrid compression. We categorize representative methods under each paradigm and evaluate them based on three criteria: \textbf{local details} (whether retains important local details), \textbf{global semantics} (whether facilitates understanding of overall context), and \textbf{inference cost} (whether reduces memory usage and inference latency).
}
\label{intro}
\vspace{-0.55cm}
\end{figure}

Context compression alleviates the difficulties of processing long contexts and reduces computational demands by selectively preserving critical information from extensive texts \citep{efficient, promptsurvey}. However, retaining sufficient information in long-context scenarios remains a substantial challenge. 
As shown in Figure \ref{model} and Appendix \ref{case}, the \textit{George Rankin} example highlights the importance of preserving both global semantics (distinguishing two individuals) and local details like names and roles (Sir George Claus Rankin, a British judge vs. Major General George James Rankin, an Australian soldier and politician).
Losing either compromises the quality of reasoning and downstream task performance.
Therefore, achieving a balance among several critical information remains a key challenge: (1) Local Detail Preservation, which requires accurately retaining important information units without introducing redundancy; (2) Global Semantic Completeness, demanding the compressed text capture the core meaning, maintain contextual coherence, and avoid omitting critical semantics; and (3) Inference Efficiency, requiring minimizing computational resources while maintaining high information density.



Current context compression research primarily focuses on hard compression and soft compression, each involving inherent trade-offs among efficiency, detail, and semantic preservation (Figure \ref{intro}). Hard compression selects natural language segments based on metrics like logits or perplexity \citep{llmlingua, selecttive, llmlingua2}, but often sacrifices fluency, coherence, and the handling of removed context \citep{promptpro, taco}, while its reliance on chunking increases time complexity \citep{nano, longllmlingua, adacomp}. Conversely, soft compression encodes text into dense latent representations for higher compression rates and scalability \citep{adapting, gist, spc}. However, this approach disrupts sequential structure, neglects local details, reduces interpretability, and complicates information tracing \citep{silver, understanding}. 
Given these limitations, a key research question arises: \textit{Can we combine the specificity of explicit tokens with the abstraction of latent representations to achieve a balance between local detail and global information retention in context compression?}

To answer the above question, we propose \textbf{Hy}brid \textbf{Co}ntext \textbf{Co}mpression (HyCo$_2$) to achieve effective information retention from both global and local perspectives. 
As shown in Figure \ref{intro} right, drawing inspiration from how humans process information from a coarse global understanding to fine local details, 
HyCo$_2$ employs a dual-level compression framework designed to retain both global semantics and local details.
\textbf{Global Compression} leverages a hybrid adapter, combining the strengths of MLPs \citep{llava}, Q-Former \citep{blip}, and Resampler \citep{flamingo}, which captures overarching contextual information through joint local and global attention mechanisms:
\textit{Local attention} segments the input context into groups and compresses each group into a single token, maintaining structural coherence and emphasizing subregions.
\textit{Global attention} utilizes learnable tokens that interact with both the instruction and the entire context to extract key global semantics.
\textbf{Local Compression} employs an auxiliary classification layer trained to identify and retain critical tokens \citep{selection}, ensuring fine-grained details necessary for accurate reasoning are preserved.
The outputs of local and global attention are then softly fused, producing a rich, instruction-aware representation that is subsequently passed to the frozen LLM.
At the same time, we find that it is challenging to train the global and local compression simultaneously.
To fully leverage HyCo$_2$’s potential, we propose pretraining the global and the local compression module using paraphrase and completion tasks respectively before instruction tuning. This alternating training strategy enables effective learning and utilization of both global and local representations. Extensive empirical studies validate the effectiveness of our HyCo$_2$. Remarkably, our approach achieves leading performance across various models on 7 datasets with significantly fewer costs, even matching the performance of the original context. Our main contributions are listed as follows:
\begin{itemize}[leftmargin=*]
    \item We propose HyCo$_2$, a hybrid context compression method for LLMs that balances hard and soft compression using a dual-level compression strategy. HyCo$_2$ effectively reduces computational costs while enabling efficient understanding of long context. 
    \item HyCo$_2$ is designed for minimal parameter updates without relying on additional compressors or external embedding models, which ensures that both the training and inference are lightweight.
    \item We propose an alternating pretraining strategy for global and local compression modules using paraphrase and completion tasks respectively to further enhance the effectiveness of HyCo$_2$. 
    \item Extensive experiments on multiple benchmarks show that HyCo$_2$ achieves superior performance compared to existing methods with significantly lower computational overhead, thereby offering valuable insights into designing effective hybrid context compression strategies for LLMs. 
\end{itemize}



\section{Related Work}

Context compression aims to reduce the input length of LLMs while preserving essential information.
Existing methods typically fall into two paradigms: \textit{hard compression} and \textit{soft compression}.

\noindent \textbf{Hard compression} acts as a filtering mechanism, reducing input length by retaining natural language tokens or paraphrasing text while aiming to preserve task-relevant information \citep{exit}. However, this approach can lead to reduced fluency, grammatical errors, and limited generalizability across LLMs with varying embedding configurations. Hard compression methods typically fall into two categories: Filtering methods such as SelectiveContext \citep{selecttive} and LLMLingua \citep{llmlingua}, use metrics like self-information or external language models to identify and remove low-utility tokens. While effective in basic scenarios, they may lack syntactic robustness and cross-model compatibility. Advanced variants like LongLLMLingua \citep{longllmlingua} and RL-driven TACO-RL \citep{taco} extend filtering to long contexts and task-specific optimization.
Paraphrasing approaches like Nano-Capsulator \citep{nano} employ fine-tuned models to generate concise rephrased prompts. While achieving condensation, these methods generally incur higher computational costs for the generation process \citep{skintern}.


\noindent \textbf{Soft compression} 
encodes context into compact continuous representations (e.g., embeddings or K-V pairs) to reduce computational costs and preserve task performance. These methods achieve higher compression rates and scalability than hard compression by discarding natural language structure. However, this often leads to substantial information loss, structural disruption, neglected local details, reduced interpretability, and complicated information tracing \citep{silver, understanding}.
Architecturally, soft compression approaches vary, including methods based on contrastive conditioning \citep{cc}, attention modification like GIST tokens \citep{gist}, recursive compression such as AutoCompressor \citep{adapting}, and memory slot encoding (e.g., ICAE \citep{icae}, 500xCompressor \citep{500xcompressor}). Inspired by multimodal techniques, some methods (e.g., xRAG \citep{xrag}, UniICL \citep{uniicl}) use MLPs to project the final token embeddings, which often results in significant information loss. Furthermore, methods like xRAG require loading additional embedding models, increasing memory overhead \citep{characterizing, promptsurvey}.
To address the limitations inherent in purely soft methods, particularly the loss of local detail and interpretability, our approach integrates hard compression. Unlike a recent soft-only hybrid method \citep{hybrid}, our method (HyCo$_2$) preserves critical local details and textual structure. By leveraging instruction-aware grouped pooling and Q-Former mechanisms for soft fusion, our approach enhances the preservation of global semantics and instruction-relevant information.
Crucially, HyCo$_2$ maintains high efficiency and scalability by introducing only a small number of additional parameters 
, avoiding the need for extra models.

\section{Methodology}

\begin{figure}[t]
\centerline{\includegraphics[width=1.0\textwidth]{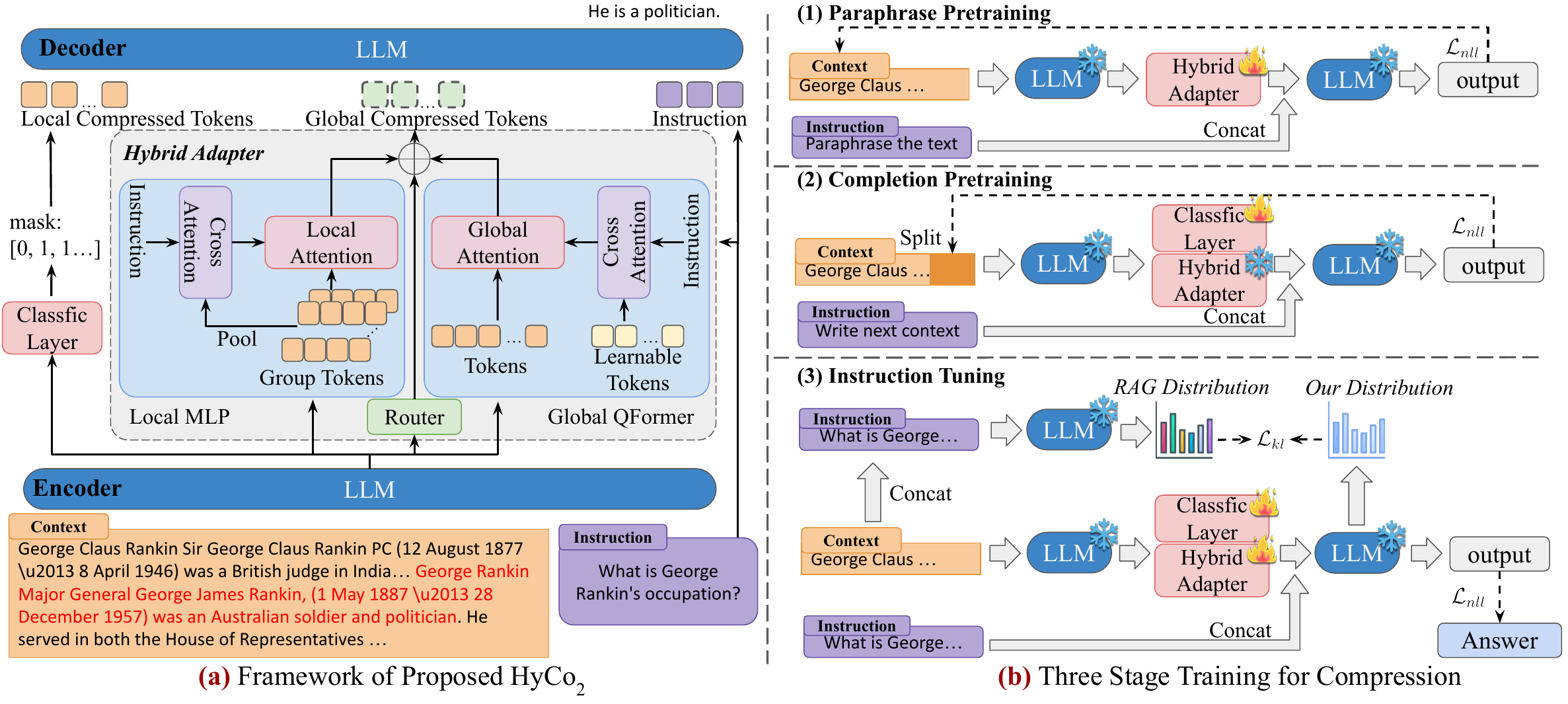}}
\caption{\textcolor{red}{(a)} \textbf{Hybrid Context Compression Framework.}
We employ a classification layer for local tokens selection and use a hybrid adapter to extract instruction-relevant representation.
Additionally, a router optimizes the global context through soft integration, thereby optimizing overall context representation.
\textcolor{red}{(b)} \textbf{Alternating Training Method.} (1) Refining the hybrid adapter with paraphrase pretraining, (2) optimizing the classification layer with completion pretraining and (3) instruction tuning for both the hybrid adapter and the classification layer.
}
\label{model}
\vspace{-0.4cm}
\end{figure}

This section begins with an overview of foundational concepts in LLM-based context compression (Section \ref{preli}). We then detail our Hybrid Context Compression framework (Section \ref{hcc}), which integrates global context refinement through a soft mixture-of-experts (MoE) mechanism, complemented by a classification layer to address hard compression of local features. Section \ref{alter} introduces an alternating training strategy to align compressed textual representations with the LLM’s semantic space. Figure \ref{model} shows the model architecture and training workflow of the proposed methodology.

\subsection{Preliminaries}
\label{preli}

Context compression aims to reduce the length of input context while preserving its functional utility in guiding LLMs to perform downstream tasks effectively. This is particularly important as the complexity of tasks increases, necessitating longer context that can lead to higher memory usage and slower inference speeds.
Formally, given a context represented as a sequence of tokens $\boldsymbol{x} = (x_1, x_2, \dots, x_N)$, where $N = |\boldsymbol{x}|$ denotes the sequence length, the objective of context compression is to identify a shorter sequence $\hat{\boldsymbol{x}}$ such that:
\begin{equation}
\label{eq1}
    \min_{\hat{\boldsymbol{x}}} \mathcal{D}(f(\cdot|\boldsymbol{x}), f(\cdot|\hat{\boldsymbol{x}})), \quad \text{s.t. } |\hat{\boldsymbol{x}}| \leq |\boldsymbol{x}|
\end{equation}
where $f(\cdot|\boldsymbol{x})$ represents the conditional distribution over the original context $\boldsymbol{x}$, $f(\cdot|\hat{\boldsymbol{x}})$ represents the conditional distribution over the compressed context $\hat{\boldsymbol{x}}$, and $\mathcal{D}$ is a divergence metric (e.g., Kullback-Leibler divergence) that quantifies the difference between the two distributions. The goal is to minimize $\mathcal{D}$, ensuring that the compressed $\hat{\boldsymbol{x}}$ retains essential information from the original $\boldsymbol{x}$. 

\subsection{Hybrid Context Compression}
\label{hcc}

Human cognition processes inputs holistically, prioritizing integrated perception before attending to granular details. Inspired by this mechanism, we propose a hybrid context compression framework that unifies hard compressed local features (capturing fine-grained textual variations) with soft gated global semantics (encoding high-level contextual understanding). 

\begin{wrapfigure}[19]{r}{0.35\textwidth}
\vspace{-0.2cm}
\centering
\includegraphics[width=0.35\textwidth]{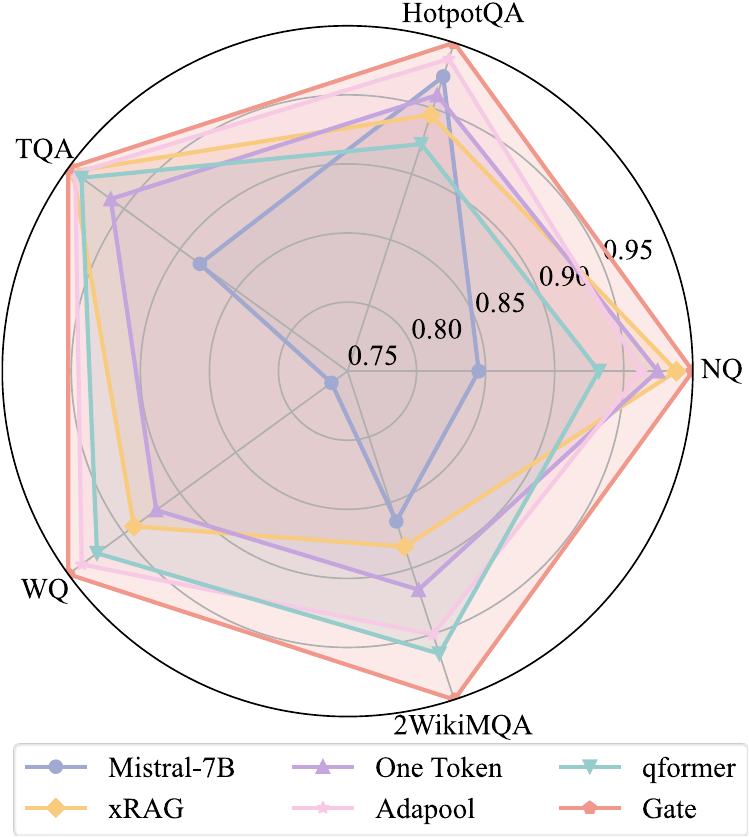}
\caption{ Significance of Soft MoE. The reported values represent the performance ratio of baselines to the best one: Gate.
}
\label{mo}
\end{wrapfigure}

\noindent \textbf{Why Soft Mixture of Experts?} Our methodology is informed by empirical insights consistent with prior multimodal research: while Query Transformer (QFormer) \footnote{In this context, the abbreviation ’QFormer’ refers to query former, where we utilize learnable query embeddings as described in previous works \citep{hybrid, beyond}, rather than employing the QFormer \citep{blip} approach.} offer superior flexibility and expressive power for contextual compression compared to multilayer perceptrons (MLPs), they demand meticulous hyperparameter optimization to match the performance of structurally simpler MLPs. 
As shown in Figure \ref{mo}, substituting MLPs (Adapool) with QFormer under fixed query tokens constraints leads to marked performance degradation across most tasks. 
This suggests that a simpler structure may facilitate more effective assimilation of compressed context by LLMs. However, in specific tasks, such as multi-document reasoning on 2WIKI, the QFormer demonstrates an advantage. Through learnable query tokens and attention mechanisms, it can dynamically prioritize task-relevant features, thereby enhancing context awareness and reasoning capabilities.
Notably, even employing a single learnable token (One Token) can yield performance comparable to the xRAG \citep{xrag}, which demonstrates that single token projection with MLPs causes severe information loss, particularly in reasoning tasks. These observations underscore the inherent limitations of relying on a single compression mechanism and motivate the investigation of hybrid approaches for more effective refinement of semantic representations.


\noindent \textbf{Soft Global Context Refinement within Hybrid Adapter.} Building on the insights from our analysis, we propose a novel method that optimizes global semantics by synergistically leveraging the strengths of MLPs and QFormer. Specifically, we employ a noisy mixture-of-experts (MoE) framework to unify these two architectural paradigms \citep{beyond}. In this framework, for feature \( \boldsymbol{V} \in \mathbb{R}^{S \times D} \) derived from Encoder (i.e., final hidden states of Encoder), where \( S \) denotes the input length and \( D \) the embedding dimension, a learned gating network \( \mathcal{G} \) dynamically determines the fusion weights for the two adapters:  
$
\mathcal{G}(\boldsymbol{V})_0 \cdot f_m(\boldsymbol{V}) + \mathcal{G}(\boldsymbol{V})_1 \cdot f_q(\boldsymbol{V})
$  
, where \( f_m(\cdot) \) and \( f_q(\cdot) \) denote the MLPs and QFormer branches, respectively. We inject learnable noise during training to mitigate the gating network’s tendency to favor a single adapter disproportionately. This is formalized with a standard normal distribution $\mathcal{N}(0, 1)$, router weight matrix $\mathbf{W}_g$ and noise weight matrix $\mathbf{W}_{\text{noise}}$:  
\begin{equation}
    \mathcal{G}(\boldsymbol{V}) = \text{Softmax}\left( \left\{ \left( \boldsymbol{V} \cdot \mathbf{W}_g \right)_i + \mathcal{N}(0, 1) \cdot \text{Softplus}\left( \boldsymbol{V} \cdot \mathbf{W}_{\text{noise}} \right)_i \right\}_{i=1}^2 \right)
\end{equation}
To enhance instruction awareness, we integrate cross-attention mechanisms with instruction embedding $\boldsymbol{C}$ into both the MLP (\( f_m(\cdot) \)) and QFormer (\( f_q(\cdot) \)) branches. For local attention in the MLP branch, we first segment the input features into $\boldsymbol{n}$ distinct groups, where $\boldsymbol{n}$ corresponds to the number of learnable tokens in the QFormer. Each group \(\boldsymbol{V}^{\boldsymbol{i}}\) (where \( 0 \leq \boldsymbol{i} < \boldsymbol{n} \)) contains \(\left\lceil S/\boldsymbol{n} \right\rceil \) tokens, which are condensed through average pooling into a single representative token \(\boldsymbol{V}^{\boldsymbol{i}}_p\) for instruction interaction. Then the local attention within each group is defined as follows:
\begin{equation}
 f_m(\boldsymbol{V}) = \bigoplus_{i=0}^{\boldsymbol{n}-1} \mathrm{MLP}(\mathrm{Attn}( \underbrace{\mathrm{CrossAttn}(\boldsymbol{V}^{\boldsymbol{i}}_p, \boldsymbol{C})}_{\mathrm{Query}}, \underbrace{\boldsymbol{V}^{\boldsymbol{i}}}_{\mathrm{Key}}, \underbrace{\boldsymbol{V}^{\boldsymbol{i}}}_{\mathrm{Value}} ))
\end{equation}
where \( \text{Attn}(\cdot) \) denotes the standard attention mechanism, parameterized by query, key, and value matrices, while \( \mathrm{CrossAttn}(\cdot) \) denotes instruction-context fusion.  
While local attention mechanisms preserve textual structure by restricting focus to localized sub-regions, this approach risks incorporating instructionally irrelevant content within partitioned regions. To mitigate this limitation, we employ the QFormer to dynamically identify and emphasize portions of the context most critical to the given instruction. Specifically, we introduce a learnable token set \( \boldsymbol{L} \in \mathbb{R}^{N_L \times D} \), where \( N_L \) denotes the token count. This token set interacts with the instruction embedding \(\boldsymbol{C}\) through cross-attention, augmented by positional embeddings \( \mathrm{Pos}(\cdot) \). The resulting global attention is computed as:  
\begin{equation}
    f_q(\boldsymbol{V}) = \mathrm{Attn}\left( \mathrm{CrossAttn}(\boldsymbol{L}, \boldsymbol{C}), \mathbf{V} + \mathrm{Pos}(\mathbf{V}), \mathbf{V} \right)
\end{equation}

\noindent \textbf{Hard Selective Local Context Mining through Classification Layer.} The information content of each token $x_i$ is quantified by a retention probability $\boldsymbol{p}_i \in [0, 1]$, with higher values indicating greater significance. Consistent with previous research \citep{selection}, we avoid designing a separate deep network for this estimation. Instead, we leverage the feature $\boldsymbol{V} = \{\boldsymbol{v}_1, \boldsymbol{v}_2, \dots, \boldsymbol{v}_n\}$, where $\boldsymbol{v}_i$ corresponds to the token $x_i$. A linear projection layer processes these feature to compute the vector of retention probabilities $\boldsymbol{p} = [\boldsymbol{p}_1, \dots, \boldsymbol{p}_n]$ via
$\boldsymbol{p} = \sigma(\mathbf{W} \boldsymbol{V} + b)$, where $\sigma$ represents the Sigmoid function, ensuring outputs lie within $[0, 1]$. $\mathbf{W}$ and $b$ are the linear layer's weight matrix and bias vector, respectively, which are learned parameters mapping feature to probabilities. Based on a target compression ratio (e.g., keeping the Top-$k\%$), tokens associated with the highest $\boldsymbol{p}_i$ values are retained. 
Furthermore, the generation of $\boldsymbol{p}$ can be integrated into a single forward pass shared with the previously described global compression strategy, thereby reducing computational overhead.


\subsection{Alternating Training Strategy}
\label{alter}
We designed a three-stage training strategy for the classification layer and hybrid adapter (Figure \ref{model} (b)), motivated by challenges in achieving optimal convergence when training both simultaneously (akin to a bilinear problem \citep{beyond}).
Stage 1: The hybrid adapter is pre-trained via a paraphrase task to reconstruct context using $\mathcal{G}(\boldsymbol{V})$ by minimizing the negative log-likelihood loss $\mathcal{L}_{\text{nll}}$.
Stage 2: With the hybrid adapter frozen, the local compression classification layer undergoes further pre-training using a completion task, also optimizing $\mathcal{L}_{\text{nll}}$.
Stage 3: Global and local compression are fine-tuned together with instruction tuning, balancing interaction for better information preservation. This involves minimizing both language modeling loss $\mathcal{L}_{\text{nll}}$ and a KL divergence term $\mathcal{L}_{\text{kl}}$ (Equation \ref{eq1}) against a teacher RAG paradigm on a hybrid open-source dataset. The final loss is the linear combination controlled by a hyperparameter: $\mathcal{L}_{\text{nll}} + \alpha\mathcal{L}_{\text{kl}}$.
We observed experimentally that single-stage training of the adapter and local compression yields inferior results, likely because the model prioritizes learning easier global features. Therefore, training the local compression components is restricted to Stage 2, enforcing a sequence of feature projection followed by local compression.
The detailed training strategy and modeling objectives are provided in Appendix \ref{training}.

\section{Experiments}
\label{main_ex}


\subsection{Experimental Setup}
\label{base}

\noindent \textbf{Datasets.} We follow the settings of \citep{xrag}, utilizing 17 datasets from reading comprehension, summarization, and open-domain QA for instruction tuning. The retrieval corpus is based on the December 2021 Wikipedia dump, with Contriever \citep{contriever} as the default retriever. By default, the instruction tuning stage uses the \textit{top-5} retrieved documents, while the downstream evaluation phase uses the \textit{top-3}. For completion pertaining (Stage 2), we use the “2023-06” snapshot from RedPajama-Data-V2 \citep{redpajama}.
We evaluate our method on 7 QA datasets, including 5 open-domain QA datasets: \textbf{NaturalQuestions (NQ)} \citep{natural}, \textbf{TriviaQA (TQA)} \citep{triviaqa}, \textbf{WebQuestions (WQ)} \citep{wq}, \textbf{PopQA (PQA)} \citep{popqa}, and \textbf{ComplexWebQuestions (CWQ)} \citep{cwq}, which cover a broad range of topics, as well as 2 multi-hop QA datasets: \textbf{HotpotQA (HQA)} \citep{hotpotqa} and \textbf{2WikiMultihopQA (2WIKI)} \citep{2wiki}, which require multi-step reasoning for answer generation. In line with prior work, we use the \textbf{Exact Match (EM)} metric to assess performance. We provide detailed information about these datasets in the Appendix \ref{apdata}.

\noindent \textbf{Implementation Details.} Evaluations of HyCo$_2$ are conducted using \textbf{LLaMA3.1-8B-Instruct} \citep{LLaMA3}, \textbf{Qwen2.5-7B-Instruct} \citep{qwen25}, and \textbf{Mistral-7B-Instruct-v0.2} \citep{Mistral7}, with the base LLM kept frozen during training. The hybrid adapter and classification layer are randomly initialized.
We set the number of query tokens ($N_L$) to 16 and the keeping ratio ($k$\%) to 10\% by default. 
We use the learning rate of 1e-4 at the pretraining stage and 2e-5 in the instruction tuning stage. 
We train 1 epoch for all stages on 8×NVIDIA A100 GPUs (80GB). 
More implementation details are in the Appendix \ref{sec:imp}.

\noindent \textbf{Baselines.} Since the LLM in our method remains frozen, the selected baselines must support plug-and-play functionality without requiring any alteration to the LLM’s parameters \citep{gist, learning}. Accordingly, we focus on three categories of baselines:
1) \colorbox{blue!10}{Uncompressed}:
\textbf{Vanilla}: Represents the original LLM, which generates answers directly without utilizing any external information.
\textbf{RAG}: Appends the top retrieved documents to the LLM's input prompts, explicitly instructing the model to reference them when generating answers.
2) \colorbox{green!10}{Hard Compression}:
\textbf{TF-IDF}: Performs topic-based discrete compression using term frequency-inverse document frequency.
\textbf{LongLLMLingua} \citep{longllmlingua} uses LLaMA2-7B-chat for token-level extraction with a 0.4 dynamic compression rate.
\textbf{LLMLingua2} \citep{llmlingua2}: A RoBERTa model trained on compressed data distilled from GPT-4.
\textbf{EXIT} \citep{exit}: Adaptively classifies and extracts contextually dependent sentences from retrieved documents.
\colorbox{yellow!25}{Soft Compression}:
\textbf{xRAG} \citep{xrag}: Uses MLPs to project the last token representation of the \textit{top-1} document.

\begin{table}[t]
  \caption{Performance comparison between our \colorbox{orange!20}{HyCo$_2$} and other methods (\colorbox{blue!10}{Uncompressed}, \colorbox{green!10}{Hard} and \colorbox{yellow!25}{Soft} compression) on seven downstream tasks. Percentages in brackets denote the relative improvement over the non-retrieval (Vanilla) setting in average performance (Avg.) and RAG setting in context length. The best results are in \textbf{bold} and the \underline{underline} indicates the dataset is IID. LLMs are frozen during the experiments and retrieved documents are set the same for different methods.}
  \label{m_table}
  \centering
  \renewcommand\arraystretch{1.05}
  \resizebox{\linewidth}{!}{
      \begin{tabular}{L{1em}lrrcccccccl}
        \toprule
        & & \textbf{Addit.} & \textbf{\# Context} &  \multicolumn{5}{c}{\textbf{Open-Domain QA} (EM $\uparrow$)} & \multicolumn{2}{c}{\textbf{Multihop QA} (EM $\uparrow$)} &   \\
        \cmidrule(r){5-9}\cmidrule(r){10-11}
        &\textbf{Methods} & \textbf{Size} $\downarrow$ & \textbf{Length} $\downarrow$   & \textbf{\underline{NQ}} & \textbf{\underline{TQA}} & \textbf{WQ} & \textbf{PQA} & \textbf{CWQ} & \textbf{HQA} & \textbf{2WIKI} & \textbf{Avg.} \\
        \midrule
        \multirow{9}*{\rotatebox{90}{Mistral-7B-Ins.-v0.2}} & \colorbox{blue!10}{Vanilla} & - & 0 \small{\textcolor{teal}{($\downarrow$\textbf{100\%})}} & 34.4 & 59.4 & 42.2 & 21.3 & 48.0 & 26.4 & 36.7 & 38.34 \small{(0.0\%)} \\ 
        & \colorbox{blue!10}{RAG} & - & 466.9 \small{(\textbf{100\%})} & 54.4 & 71.3 & 45.1 & 67.0 & 45.7 & 29.5 & 40.6 & 50.51 \small{\textcolor{purple}{($\uparrow$ 31.7\%)}} \\ 
        \cdashline{2-12}
        & \colorbox{green!10}{TF-IDF} & - & 64 \small{\textcolor{teal}{($\downarrow$ 86.3\%)}} & 34.4 & 60.6 & 38.8 & 30.7 & 43.3 & 23.0 & 39.6 & 38.63 \small{\textcolor{purple}{($\uparrow$ 0.8\%)}} \\ 
        & \colorbox{green!10}{LongLLMLingua} \citep{longllmlingua} & 7B & 131.2 \small{\textcolor{teal}{($\downarrow$ 71.9\%)}} & 39.5 & 64.3 & 39.3 & 44.3 & 49.0 & 24.9 & 39.0 & 42.90 \small{\textcolor{purple}{($\uparrow$ 11.9\%)}} \\ 
        & \colorbox{green!10}{LLMLingua2} \citep{llmlingua2} & 561M & 114.2 \small{\textcolor{teal}{($\downarrow$ 75.5\%)}} & 38.1 & 62.5 & 41.1 & 43.7 & 45.0 & 25.5 & 38.9 & 42.11 \small{\textcolor{purple}{($\uparrow$ 9.8\%)}} \\ 
        & \colorbox{green!10}{EXIT} \citep{exit} & 4B & 83.7 \small{\textcolor{teal}{($\downarrow$ 82.0\%)}} & \textbf{41.9} & 65.4 & 43.0 & \textbf{47.3} & 49.0 & 27.2 & 39.9 & 44.81 \small{\textcolor{purple}{($\uparrow$ 16.8\%)}} \\ 
        & \colorbox{yellow!20}{xRAG} \citep{xrag} & 7B + 35M & \textbf{3} \small{\textcolor{teal}{($\downarrow$ 99.4\%)}} & 37.2 & 65.5 & 43.4 & 39.3 & 47.7 & 22.0 & 25.9 & 40.14 \small{\textcolor{purple}{($\uparrow$ 4.7\%)}} \\ 
        & \colorbox{orange!20}{\textbf{HyCo$_2$}} (\textit{ours}) & \textbf{168M} & 50.7 \small{\textcolor{teal}{($\downarrow$ 89.1\%)}} & 39.6 & \textbf{66.0} & \textbf{45.4} & 45.7 & \textbf{50.3} & \textbf{27.5} & \textbf{40.2} & \textbf{44.96} \small{\textcolor{purple}{($\uparrow$ 17.3\%)}} \\
        
        \midrule
        \multirow{9}*{\rotatebox{90}{LLaMA-3.1-8B-Ins.}} & \colorbox{blue!10}{Vanilla} & - & 0 \small{\textcolor{teal}{($\downarrow$\textbf{100\%})}} & 38.0 & 67.0 & 50.6 & 33.0 & 49.0 & 27.7 & 31.9 & 42.46 \small{(0.0\%)} \\ 
        & \colorbox{blue!10}{RAG} & - & 466.9 \small{(\textbf{100\%})} & 52.6 & 71.0 & 40.4 & 60.3 & 40.0 & 27.3 & 34.0 & 46.51 \small{\textcolor{purple}{($\uparrow$ 9.5\%)}} \\ 
        \cdashline{2-12}
        & \colorbox{green!10}{TF-IDF} & - & 64 \small{\textcolor{teal}{($\downarrow$ 86.3\%)}} & 37.0 & 64.7 & 35.4 & 27.0 & 41.3 & 23.0 & 31.3 & 37.10 \small{\textcolor{teal}{($\downarrow$ 12.6\%)}} \\ 
        & \colorbox{green!10}{LongLLMLingua} \citep{longllmlingua} & 7B & 131.2 \small{\textcolor{teal}{($\downarrow$ 71.9\%)}} & 38.1 & 66.4 & 34.3 & 40.3 & 49.0 & 25.7 & 32.4 & 40.89 \small{\textcolor{teal}{($\downarrow$ 3.7\%)}} \\ 
        & \colorbox{green!10}{LLMLingua2} \citep{llmlingua2} & 561M & 114.2 \small{\textcolor{teal}{($\downarrow$ 75.5\%)}} & 37.4 & 65.2 & 35.8 & 39.7 & 42.0 & 24.9 & 31.5 & 39.50 \small{\textcolor{teal}{($\downarrow$ 7.0\%)}} \\ 
        & \colorbox{green!10}{EXIT} \citep{exit} & 4B & 83.7 \small{\textcolor{teal}{($\downarrow$ 82.0\%)}} & \textbf{41.5} & 66.5 & 40.1 & \textbf{47.3} & 48.7 & 29.9 & 33.1 & 43.87 \small{\textcolor{purple}{($\uparrow$ 3.3\%)}} \\ 
        & \colorbox{yellow!20}{xRAG} \citep{xrag} & 7B + 35M & \textbf{3} \small{\textcolor{teal}{($\downarrow$ 99.4\%)}} & 35.6 & 64.8 & 40.0 & 34.7 & 49.0 & 24.1 & 28.1 & 39.47 \small{\textcolor{teal}{($\downarrow$ 7.0\%)}} \\ 
        & \colorbox{orange!20}{\textbf{HyCo$_2$}} (\textit{ours}) & \textbf{168M} & 52.1 \small{\textcolor{teal}{($\downarrow$ 88.8\%)}} & 39.3 & \textbf{67.1} & \textbf{40.8} & 46.7 & \textbf{49.7} & \textbf{30.5} & \textbf{33.6} & \textbf{43.96} \small{\textcolor{purple}{($\uparrow$ 3.5\%)}} \\ 

        \midrule
        \multirow{9}*{\rotatebox{90}{Qwen-2.5-7B-Ins.}} & \colorbox{blue!10}{Vanilla} & - & 0 \small{\textcolor{teal}{($\downarrow$\textbf{100\%})}} & 29.6 & 55.1 & 39.1 & 23.7 & 44.7 & 25.5 & 31.2 & 35.56 \small{(0.0\%)} \\ 
        & \colorbox{blue!10}{RAG} & - & 466.9 \small{(\textbf{100\%})} & 51.9 & 69.6 & 40.9 & 56.0 & 35.7 & 21.3 & 35.5 & 44.41 \small{\textcolor{purple}{($\uparrow$ 24.9\%)}} \\ 
        \cdashline{2-12}
        & \colorbox{green!10}{TF-IDF} & - & 64 \small{\textcolor{teal}{($\downarrow$ 86.3\%)}} & 28.9 & 56.2 & 35.3 & 11.7 & 37.3 & 20.0 & 31.8 & 31.60 \small{\textcolor{teal}{($\downarrow$ 11.1\%)}} \\ 
        & \colorbox{green!10}{LongLLMLingua} \citep{longllmlingua} & 7B & 131.2 \small{\textcolor{teal}{($\downarrow$ 71.9\%)}} & 33.4 & 59.8 & 35.3 & 43.7 & 38.7 & 21.3 & 31.7 & 37.70 \small{\textcolor{purple}{($\uparrow$ 6.0\%)}} \\ 
        & \colorbox{green!10}{LLMLingua2} \citep{llmlingua2} & 561M & 114.2 \small{\textcolor{teal}{($\downarrow$ 75.5\%)}} & 30.9 & 55.6 & 34.2 & 12.7 & 35.0 & 20.2 & 31.2 & 31.40 \small{\textcolor{teal}{($\downarrow$ 11.7\%)}} \\ 
        & \colorbox{green!10}{EXIT} \citep{exit} & 4B & 83.7 \small{\textcolor{teal}{($\downarrow$ 82.0\%)}} & \textbf{37.2} & 59.4 & 40.3 & \textbf{51.7} & 45.3 & \textbf{26.7} & 32.7 & 41.90 \small{\textcolor{purple}{($\uparrow$ 17.8\%)}} \\ 
        & \colorbox{yellow!20}{xRAG} \citep{xrag} & 7B + 35M & \textbf{3} \small{\textcolor{teal}{($\downarrow$ 99.4\%)}} & 27.9 & 53.7 & 39.7 & 23.7 & 46.0 & 23.1 & 27.9 & 34.57 \small{\textcolor{teal}{($\downarrow$ 2.8\%)}}  \\ 
        & \colorbox{orange!20}{\textbf{HyCo$_2$}} (\textit{ours}) & \textbf{168M} & 53.4 \small{\textcolor{teal}{($\downarrow$ 88.6\%)}} & 34.6 & \textbf{60.2} & \textbf{43.1} & 50.7 & \textbf{46.3} & 26.2 & \textbf{33.8} & \textbf{42.11} \small{\textcolor{purple}{($\uparrow$ 18.4\%)}} \\ 
        \bottomrule
      \end{tabular}
      }
\end{table}

\subsection{Main Results}
\label{main}

We present a comprehensive performance comparison between our proposed method HyCo$_2$ and other state-of-the-art (SOTA) techniques across 7 downstream tasks in Table \ref{m_table}. The RAG baseline, which utilizes full retrieved context without compression, significantly improves the average EM compared to the Vanilla non-retrieval setting across all LLMs
(e.g., achieving a 31.7\% relative improvement for Mistral-7B, 9.5\% for LLaMA3.1-8B and 24.9\% for Qwen2.5-7B).
Among the compression methods evaluated, our proposed HyCo$_2$ consistently achieves the highest average EM score across all three language models, demonstrating superior effectiveness in retaining relevant information compared to other techniques like EXIT and xRAG. 
Notably, HyCo$_2$ requires only 168M parameters for the additional model components during inference (excluding the reader LLM), significantly lower than xRAG's 7B and EXIT's 4B. HyCo$_2$ also drastically reduces token usage by an average reduction of 88.8\% while maintaining strong performance. In some instances, HyCo$_2$ either matches or exceeds the performance of the uncompressed method.
Specifically, with Mistral-7B, HyCo$_2$ achieves an average EM of 44.96, outperforming EXIT by 0.7\% while using 7.1\% fewer tokens. For datasets like WQ and CWQ, HyCo$_2$ surpasses the uncompressed RAG 0.7\% and 10\%, saving 89.1\% of tokens. Similar trends are observed with LLaMA3.1 and Qwen2.5.

Our experiments also reveal that for more powerful modern models, such as LLaMA3.1 and Qwen2.5, RAG underperforms compared to vanilla LLMs on certain document understanding tasks (e.g., WQ and CWQ) and multi-hop document reasoning tasks (e.g., HQA). This may be due to these tasks relying heavily on Wikipedia, whose knowledge has already been extensively absorbed during the LLM’s pretraining phase, leading to potential conflicts in knowledge. Moreover, some of the retrieved documents may contain outdated or redundant information, which could further reduce performance. This hypothesis is reinforced by the fact that most compression methods outperform RAG.
Additionally, HyCo$_2$ addresses the issue of poor multi-document reasoning performance (e.g., HQA and 2WIKI) observed in xRAG’s single-token soft compression approach \citep{xrag}.

\begin{figure}[t]
\centerline{\includegraphics[width=1.0\textwidth]{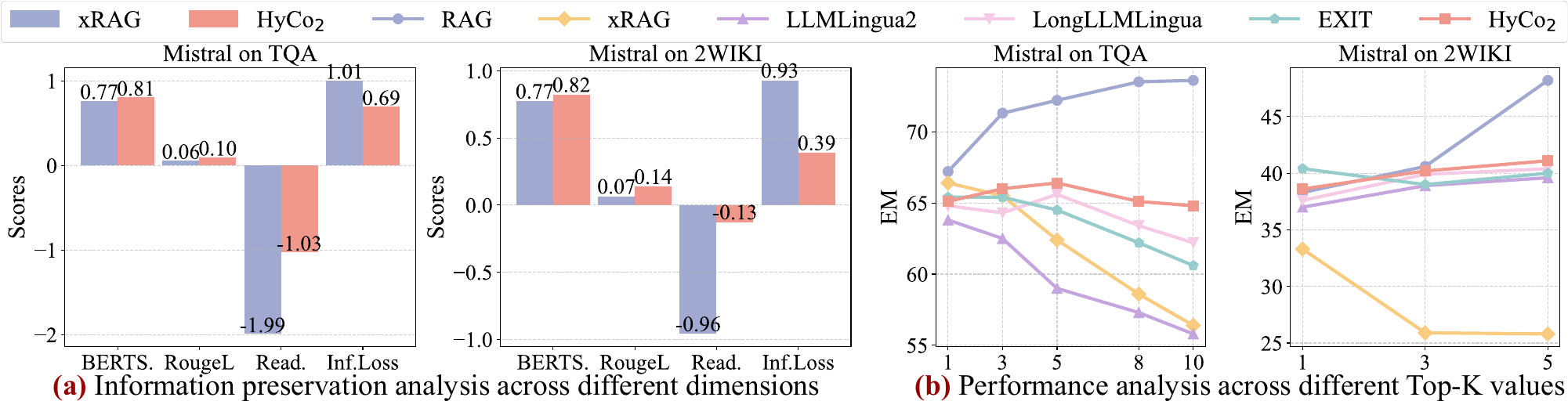}}
\caption{We employ Mistral-7B to investigate two aspects: \textcolor{red}{(a)} a four-dimensional comparison of information preservation between HyCo$_2$ and xRAG following context compression and reconstruction, and \textcolor{red}{(b)} the performance trends of various compression methods as context length increases. BERTScore measures semantic similarity, Information Loss measures the entropy value of discarded information, while Readability and ROUGE-L evaluate the quality of the reconstructed context.
}
\label{tt}
\end{figure}

\subsection{Analysis}
\label{analysis}

\noindent \textbf{Information Preservation.} To evaluate the information preservation capabilities, we prompt the target LLM to reconstruct the original context from the compressed representations (prompts refer to the Appendix \ref{apprompt}). This evaluation focuses specifically on xRAG, excluding hard compression methods, as the latter do not introduce new content and are inherently fully interpretable.
We use four metrics BERTScore, Information Loss, ROUGE, and Readability for assessment. Detailed metrics calculations are provided in Appendix \ref{apmetric}. As Figure \ref{tt} (a) illustrates, HyCo$_2$ demonstrates superior reconstruction performance compared to xRAG on TQA and 2Wiki. Specifically, we observed an average BERTScore F1 improvement of 0.05, 0.5 lower information loss, and higher scores for both readability and ROUGE-L, which demonstrates that HyCo$_2$, through its combination of global and local mechanisms, effectively retains more information and preserves critical details.

\noindent \textbf{Robustness.} To assess the robustness and effectiveness of HyCo$_2$ in handling longer texts, we gradually increase the number of retrieved documents ($K \in \{1, 3, 5, 8, 10\}$), as shown in Figure \ref{tt} (b). When $K \leq 5$ (i.e., text length less than 1k), HyCo$_2$ performs as steadily as the RAG baseline, consistently improving EM scores. In contrast, other compression methods begin to show performance degradation when $K \geq 3$. This trend is particularly evident for xRAG, which exhibits optimal performance only with a top-1 document, consistent with the results and settings reported by \citep{xrag}.
While all compression methods experience performance decline at higher $K$, HyCo$_2$'s degradation is notably slower compared to others. For instance, at $K = 10$, HyCo$_2$'s EM score drops by only 1.2 points, demonstrating superior robustness in handling longer contexts. This underscores the inherent challenges current compression methods face with longer texts, where substantial information loss persists and significant room for improvement remains.

\noindent \textbf{Efficiency and Memory.}
We utilize \textit{Torch Profiler}\footnote{\url{https://docs.pytorch.org/docs/stable/profiler.html}} to evaluate the efficiency across different methods on various datasets, measuring CPU time (s), CUDA time (s), computations (GFLOPs), and peak GPU memory usage (GB). All experiments are conducted using Mistral-7B and LLaMA3.1-8B in BFloat16 inference mode on a single A100 GPU, with a batch size of 1 and a fixed output length of 30.
As shown in Table \ref{tab:profile}, HyCo$_2$ achieves the best performance in terms of CPU time (0.572 s) and CUDA time (0.187 s). It also attains the lowest peak memory usage (14.56 GB), saving approximately 50\% GPU memory compared to xRAG, which is consistent with the additional memory overhead from xRAG’s embedding model.
In terms of GFLOPs (312.73), HyCo$_2$ outperforms xRAG and LLMLingua2, while remaining significantly more efficient than EXIT. Notably, although xRAG has the lowest GFLOPs, it exhibited the highest memory consumption. In contrast, EXIT incurres the highest computational and time costs among all methods.


\begin{table}[t] 
  \centering 

  \begin{minipage}[t]{0.49\linewidth} 
    \centering 
    \caption{Comparison of context compression methods about efficiency and memory usage.} 
    \label{tab:profile} 
    \renewcommand{\arraystretch}{1.05}
    \resizebox{\linewidth}{!}{ 
      \begin{tabular}{lC{5em}C{5em}C{5em}C{5em}}
        \toprule
        \textbf{Method} & \textbf{CPU Time (s)} & \textbf{CUDA Time (s)} & \textbf{GFLOPs} & \textbf{Peak Mem. (GB)} \\
        \midrule
        \multicolumn{5}{l}{\textcolor{gray}{\textit{Mistral-7B-Instruct-v0.2 on TQA}}} \\
        \colorbox{yellow!20}{xRAG} & 0.716 & 0.249 & \textbf{253.25} & 27.05 \\
        \colorbox{green!10}{LLMLingua2} & 1.037 & 0.418 & 264.77 & 16.60 \\
        \colorbox{green!10}{EXIT} & 2.495 & 0.820 & 1624.37 & 20.43 \\
        \colorbox{orange!20}{\textbf{HyCo$_2$}} (\textit{ours})& \textbf{0.572} & \textbf{0.187} & 312.73 & \textbf{14.56} \\
        \midrule
        \multicolumn{5}{l}{\textcolor{gray}{\textit{Mistral-7B-Instruct-v0.2 on 2WIKI}}} \\
        \colorbox{yellow!20}{xRAG} & 0.787& 0.252& \textbf{181.89} & 27.06 \\
        \colorbox{green!10}{LLMLingua2} & 1.031& 0.408& 192.58 & 16.60 \\
        \colorbox{green!10}{EXIT} & 1.639& 0.626& 1142.99 & 20.41 \\
        \colorbox{orange!20}{\textbf{HyCo$_2$}} (\textit{ours}) & \textbf{0.672}& \textbf{0.197}& 228.50 & \textbf{14.78} \\
        \midrule
        \multicolumn{5}{l}{\textcolor{gray}{\textit{LLaMA3.1-8B-Instruct on TQA}}} \\
        \colorbox{yellow!20}{xRAG} & 0.591 & 0.248 & 251.99 & 28.52 \\
        \colorbox{green!10}{LLMLingua2} & 0.656 & 0.178 & \textbf{242.95} & 18.50 \\
        \colorbox{green!10}{EXIT} & 1.456 & 0.665 & 1602.54 & 21.90 \\
        \colorbox{orange!20}{\textbf{HyCo$_2$}} (\textit{ours})& \textbf{0.324} & \textbf{0.136} & 288.05 & \textbf{16.92} \\
        \midrule
        \multicolumn{5}{l}{\textcolor{gray}{\textit{LLaMA3.1-8B-Instruct on 2WIKI}}} \\
        \colorbox{yellow!20}{xRAG} & 0.575 & 0.228 & \textbf{180.02} & 28.53 \\
        \colorbox{green!10}{LLMLingua2} & 0.854 & 0.234 & 188.04 & 18.80 \\
        \colorbox{green!10}{EXIT} & 0.916 & 0.395 & 962.47 & 21.88 \\
        \colorbox{orange!20}{\textbf{HyCo$_2$}} (\textit{ours}) & \textbf{0.334}& \textbf{0.126}& 211.53 & \textbf{17.38} \\
        \bottomrule
      \end{tabular}
    }
  \end{minipage}
  \hfill 
  \begin{minipage}[t]{0.49\linewidth} 
    \centering 
    
    \caption{Results of Ablation Studies. The row with a \colorbox{gray!10}{gray background} indicates our default setting. The backbone model is Mistral-7B.} 
    \label{abl} 
    \renewcommand\arraystretch{1.05}
    \resizebox{\linewidth}{!}{ 
      \begin{tabular}{lllll}
        \toprule
        \textbf{Method} & \textbf{NQ} & \textbf{TQA} & \textbf{HQA} & \textbf{2WIKI} \\
        \midrule
        \rowcolor{gray!10} HyCo$_2$ & 39.6 & 66.0 & 27.5 & 40.2 \\
        ~~w/o Ins. & 38.8 \footnotesize{(\textcolor{teal}{-0.8})} & 65.5 \footnotesize{(\textcolor{teal}{-0.5})} & 26.1 \footnotesize{(\textcolor{teal}{-1.4})} & 38.6 \footnotesize{(\textcolor{teal}{-1.6})} \\
        ~~w/o $\mathcal{L}_{\text{nll}}$ & 37.7 \footnotesize{(\textcolor{teal}{-1.9})} & 63.9 \footnotesize{(\textcolor{teal}{-2.1})} & 26.7 \footnotesize{(\textcolor{teal}{-0.8})} & 41.4 \footnotesize{(\textcolor{purple}{+1.2})} \\
        ~~w/o $\mathcal{L}_{\text{kl}}$ & 35.2 \footnotesize{(\textcolor{teal}{-4.4})} & 62.6 \footnotesize{(\textcolor{teal}{-3.4})} & 26.4 \footnotesize{(\textcolor{teal}{-1.1})} & 38.8 \footnotesize{(\textcolor{teal}{-1.4})} \\
        ~~w/o Pretrain & 34.2 \footnotesize{(\textcolor{teal}{-5.4})} & 59.4 \footnotesize{(\textcolor{teal}{-6.6})} & 25.0 \footnotesize{(\textcolor{teal}{-2.5})} & 38.2 \footnotesize{(\textcolor{teal}{-2.0})} \\
        ~~w/o Finetune & 33.1 \footnotesize{(\textcolor{teal}{-6.5})} & 60.7 \footnotesize{(\textcolor{teal}{-5.3})} & 25.6 \footnotesize{(\textcolor{teal}{-1.9})} & 39.4 \footnotesize{(\textcolor{teal}{-0.8})} \\
        \midrule
        \multicolumn{5}{l}{\textcolor{gray}{\textit{Query Type}}} \\
        One Token & 33.5 \footnotesize{(\textcolor{teal}{-6.1})} & 60.0 \footnotesize{(\textcolor{teal}{-6.0})} & 25.4 \footnotesize{(\textcolor{teal}{-2.1})} & 37.1 \footnotesize{(\textcolor{teal}{-3.1})} \\
        AdaPool & 36.4 \footnotesize{(\textcolor{teal}{-3.2})} & 63.0 \footnotesize{(\textcolor{teal}{-3.0})} & 28.0 \footnotesize{(\textcolor{purple}{+0.5})} & 38.9 \footnotesize{(\textcolor{teal}{-1.3})} \\
          QFormer & 34.7 \footnotesize{(\textcolor{teal}{-4.9})} & 63.9 \footnotesize{(\textcolor{teal}{-2.1})} & 26.8 \footnotesize{(\textcolor{teal}{-0.7})} & 37.7 \footnotesize{(\textcolor{teal}{-2.5})} \\
        \rowcolor{gray!10} Hybrid & 39.6 & 66.0 & 27.5 & 40.2 \\
        \midrule
        \multicolumn{5}{l}{\textcolor{gray}{\textit{Training strategies}}} \\
        E2E  & 36.8 \footnotesize{(\textcolor{teal}{-2.8})} & 62.8 \footnotesize{(\textcolor{teal}{-3.2})} & 26.4 \footnotesize{(\textcolor{teal}{-1.1})} & 38.3 \footnotesize{(\textcolor{teal}{-1.9})} \\
        ~~w/o Stage 2 & 36.3 \footnotesize{(\textcolor{teal}{-3.3})} & 62.0 \footnotesize{(\textcolor{teal}{-4.0})} & 25.6 \footnotesize{(\textcolor{teal}{-1.9})} & 37.8 \footnotesize{(\textcolor{teal}{-2.4})} \\
        ~~w/o Global & 29.7 \footnotesize{(\textcolor{teal}{-9.9})} & 55.7 \footnotesize{(\textcolor{teal}{-10.3})} & 22.4 \footnotesize{(\textcolor{teal}{-5.1})} & 35.0 \footnotesize{(\textcolor{teal}{-5.2})}\\
        ~~w/o Local & 33.6 \footnotesize{(\textcolor{teal}{-6.0})} & 60.5 \footnotesize{(\textcolor{teal}{-5.5})} & 24.8 \footnotesize{(\textcolor{teal}{-2.7})} & 37.9 \footnotesize{(\textcolor{teal}{-2.3})} \\
        \rowcolor{gray!10} Alternating & 39.6 & 66.0 & 27.5 & 40.2 \\
        ~~w/o Stage 2 & 37.8 \footnotesize{(\textcolor{teal}{-1.8})} & 64.1 \footnotesize{(\textcolor{teal}{-1.9})} & 27.1 \footnotesize{(\textcolor{teal}{-0.4})} & 39.3 \footnotesize{(\textcolor{teal}{-0.9})}\\
        ~~w/o Global & 33.2 \footnotesize{(\textcolor{teal}{-6.4})} & 58.8 \footnotesize{(\textcolor{teal}{-7.2})} & 24.7 \footnotesize{(\textcolor{teal}{-2.8})} & 37.3 \footnotesize{(\textcolor{teal}{-2.9})} \\
        ~~w/o Local & 35.4 \footnotesize{(\textcolor{teal}{-4.2})} & 63.6 \footnotesize{(\textcolor{teal}{-2.4})} & 26.6 \footnotesize{(\textcolor{teal}{-0.9})} & 38.9 \footnotesize{(\textcolor{teal}{-1.3})} \\
        \bottomrule
      \end{tabular}
    }
  \end{minipage}

\end{table}

\subsection{Ablation Studies}
\label{ab_s}

\noindent \textbf{Components Analysis.} Table \ref{abl} presents a comprehensive analysis of the effectiveness of various components within HyCo$_2$. 
Removing the instruction-conditioned cross-attention leads to a notable drop in performance, highlighting that instructions provide valuable guidance for the compressor to identify key information for QA. Regarding loss functions, $\mathcal{L}_{\text{kl}}$ (self-distillation) outperforms $\mathcal{L}_{\text{nll}}$ (language modeling), as it better aligns the compressor with richer teacher representations and facilitates the learning of more salient features. Additionally, both the pretraining and instruction tuning stages are essential, each contributing substantially to overall performance and validating the effectiveness of the proposed training strategy.

\noindent \textbf{Effects of Hybrid Adapter for Global Compression.} Learnable queries are commonly used in query-based Transformers to extract salient information, whereas pooling-based projections aim to uniformly preserve information across input segments. We compared HyCo$_2$ with baseline variants incorporating these representation strategies: learnable query tokens (QFormer), pooling projection (AdaPool), and a single learnable token (One Token). The results are shown in Table \ref{abl} (Query Type section).
Compared to a single learnable token, pooling projection demonstrates superior capacity for information retention and downstream inference. Although QFormer offers a theoretical advantage in facilitating instruction interaction, its practical performance was suboptimal. Our results suggest that combining pooling and learnable queries leads to further performance gains \citep{beyond}.

\noindent \textbf{Impact of Alternating Training.} 
We further investigate the impact of the alternating training strategy on model performance. 
We first compare the alternating training strategy against direct end-to-end (E2E) training, observing a notable average performance drop of 2\%. Building on this, omitting Stage 2 pretraining leads to further performance degradation, indicating that pretraining the local compression module is essential for learning token importance effectively. To better understand this discrepancy, we conduct ablation experiments isolating global and local compression components. Training only the local compression module results in poor performance under both E2E and alternating settings, likely due to the severe semantic loss caused by retaining only 10\% of the context. In contrast, using only the global compression module yielded relatively better results, underscoring the importance of capturing global semantics during compression. 
These findings collectively highlight the necessity and effectiveness of the alternating training strategy.

\section{Conclusion}

In this paper, we introduce HyCo$_2$ (Hybrid Context Compression), a novel approach for balancing local and global information retention in large language models (LLMs). HyCo$_2$ addresses the significant challenges of long-context inference, such as computational inefficiency and redundant processing of extended input sequences. By integrating both hard compression (retaining fine-grained local details) and soft compression (capturing high-level global semantics), HyCo$_2$ achieves a harmonious trade-off between preserving instruction-relevant content and reducing token consumption. Moreover, we use an alternating training strategy that pretrains the global and local compression modules using paraphrasing and completion tasks, respectively, followed by instruction tuning to align with downstream tasks. Our experimental results demonstrate that HyCo$_2$ significantly enhances performance across various knowledge-intensive tasks, including open-domain question answering and multi-hop reasoning. 
HyCo$_2$ represents a significant step forward in context compression for LLMs, offering a hybrid, lightweight, efficient, and effective solution for long-text reasoning. 

\bibliography{custom}
\bibliographystyle{plain}


\appendix

\section{Case Study for Critical Interplay between Local Detail Preservation and Global Semantic Completeness}
\label{case}

Consider a document containing the following entries:

\begin{table*}[ht!]\centering
\begin{minipage}{0.99\columnwidth}\vspace{0mm}    \centering
\begin{tcolorbox} 
    \centering
     \hspace{-6mm}
\setlength{\itemsep}{2pt}
"\colorbox{green!40}{George Claus Rankin Sir George Claus Rankin PC} (12 August 1877 – 8 April 1946) was a British judge in India…"

"\colorbox{red!40}{George Rankin Major General George James Rankin}, (1 May 1887 – 28 December 1957) was an Australian soldier and politician. He served in both the House of Representatives…"

Question: "What is \colorbox{red!40}{George Rankin}'s occupation?"
\end{tcolorbox}
\vspace{-2mm}
    \label{table:case}
\end{minipage}
\end{table*}

\textbf{Global Semantic Completeness} is essential for accurate query interpretation in this context. It entails recognizing that the document discusses \textbf{two distinct individuals} named \textit{George Rankin}, rather than a single person. A compression method that conflates these entities or represents only the first instance fails to preserve the document’s overarching semantic structure. Specifically, it would neglect the ambiguity inherent in the term "George Rankin," omitting the fact that multiple, disambiguated profiles are present. The compressed representation must therefore retain the \textbf{core semantic meaning}: that "George Rankin" refers to more than one person, each associated with a unique set of attributes.

\textbf{Local Detail Preservation}, by contrast, concerns the retention of fine-grained, entity-specific information. For \textit{Sir George Claus Rankin}, this includes his full name, honorific title ("Sir"), professional role ("British judge in India"), and lifespan. For \textit{Major General George James Rankin}, the critical local details include his full name, military rank ("Major General"), professional roles ("Australian soldier and politician"), and service record. If the compression process omits these elements, such as the occupations or titles, it undermines the factual integrity of the representation, even if the presence of multiple entities is correctly preserved.

Accordingly, an effective compression method must satisfy both criteria. It must maintain global semantic completeness by encoding the presence of multiple individuals named "George Rankin," and simultaneously ensure local detail preservation by retaining the specific identifiers that distinguish them. A compressed output that enables a system to generate the response, "The document refers to two individuals: \textit{Sir George Claus Rankin, a British judge in India, and Major General George James Rankin, an Australian soldier and politician}," would exemplify successful integration of these principles. This case underscores that failure in either global disambiguation or local specificity significantly compromises the utility of compressed representations for downstream reasoning and information retrieval tasks.

\section{Alternating Training Strategy}
\label{training}

\subsection{Paraphrase Pretraining}

In Stage 1, the objective is to train the hybrid compressor to align the soft-gated global token with the original context $\boldsymbol{x}$'s global semantics. Specifically, the LLM utilizes natural language instructions $\Xmat_{\texttt{paraphrase}}$\footnote{To maintain diversity, we sample from an instruction pool, which could be found in Appendix~\ref{prompt_train}.} to generate context, aiming to reconstruct the original context. The optimization objective is defined by the following formula:
\begin{equation}
    \mathcal{L}_{\text{nll}} = -\sum_{i=1}\text{log}\ p_{\bm{\phi}}(x_i|\mathcal{G}(\bm{\mathcal{F}_{\phi}}(\boldsymbol{x})),\Xmat_{\texttt{paraphrase}},x_{<i})
\end{equation}
where $p_{\bm{\phi}}$ is given by the softmax distribution of LLM $\bm{\mathcal{F}_{\phi}}$, $\bm{\mathcal{F}_{\phi}}(\boldsymbol{x})$ is the context feature encoded by Encoder (LLM itself), $\mathcal{G}$ is a learned gating network and $x_{<i}$ denotes the context token before current prediction token $x_{i}$, achieved by casual attention mask in auto-regressive LMs.

\subsection{Completion Pretraining}

In Stage 2, the context $\boldsymbol{x}$ from RedPajama-Data-V2 is randomly partitioned into two segments: \(a\) and \(b\). Segment \(a\) functions as the context, while segment \(b\) is the target for prediction. By minimizing the negative log-likelihood $\mathcal{L}_{\text{nll}}$ of predicting segment \(b\) given the compressed context of \(a\) (formed using the local classification layer), the model is trained to preserve the key information from context \(a\) necessary to generate \(b\) using instructions $\Xmat_{\texttt{completion}}$. The optimization objective is:
\begin{equation}
    \mathcal{L}_{\text{nll}} = -\sum_{i=1}\text{log}\ p_{\bm{\phi}}(b_i|\mathcal{H}(\bm{\mathcal{F}_{\phi}}(a)), \mathcal{G}(\bm{\mathcal{F}_{\phi}}(a)),\Xmat_{\texttt{completion}}, b_{<i})
\end{equation}
where $\mathcal{H}$ is the local classification layer for keeping top $k$\% tokens.

\subsection{Instruction Tuning}

In Stage 3, we utilize triplets (\(\boldsymbol{q}, \boldsymbol{x}, \boldsymbol{y}\)) where \(\boldsymbol{q}\) is the question, \(\boldsymbol{x}\) is the context (retrieved documents or long input), and \(\boldsymbol{y}\) is the output answer. On one hand, we employ a language modeling objective, consistent with the first two stages, to train the model to generate the correct output \(\boldsymbol{y}\) based on task-specific instructions and the provided context \(\boldsymbol{x}\):
\begin{equation}
    \mathcal{L}_{\text{nll}} = -\sum_{i=1}\text{log}\ p_{\bm{\phi}}(\boldsymbol{y}_i|\mathcal{H}(\bm{\mathcal{F}_{\phi}}(\boldsymbol{x})),\mathcal{G}(\bm{\mathcal{F}_{\phi}}(\boldsymbol{x})), \boldsymbol{q},\boldsymbol{y}_{<i})
\end{equation}

On the other hand, we incorporate self-distillation \citep{xrag}, treating the RAG model as the teacher and HyCo$_2$ as the student to transfer knowledge. This process trains HyCo$_2$ to simulate the RAG model's proficiency in handling complete, uncompressed documents, thereby facilitating the learning of more effective compressed representations. Specifically, for the base language model $\bm{\mathcal{F}_{\phi}}$, which receives either the uncompressed context $\boldsymbol{x}$ (from the teacher RAG model) or the compressed representation $(\mathcal{H}(\bm{\mathcal{F}_{\phi}}(\boldsymbol{x})),\mathcal{G}(\bm{\mathcal{F}_{\phi}}(\boldsymbol{x})))$ (from HyCo$_2$), the objective is to minimize the divergence between the two resulting output distributions. This divergence is measured using the Kullback-Leibler (KL) divergence:
\begin{equation}
    \mathcal{L}_{\text{kl}} = \mathcal{D}_{\textrm{KL}}(p_{\bm{\phi}}(\boldsymbol{y}|\boldsymbol{x},\boldsymbol{q})\ ||\ p_{\bm{\phi}}(\boldsymbol{y}|\mathcal{H}(\bm{\mathcal{F}_{\phi}}(\boldsymbol{x})),\mathcal{G}(\bm{\mathcal{F}_{\phi}}(\boldsymbol{x})),\boldsymbol{q}))
\end{equation}
The final loss is the linear combination controlled by a hyperparameter: $\mathcal{L}_{\text{nll}} + \alpha\mathcal{L}_{\text{kl}}$.

\section{Experimantal Settings}
\label{experiment}

\subsection{Datasets}
\label{apdata}

\subsubsection{Details for Pretraining Dataset}

For Paraphrase Pretraining, we construct training instances derived from the retrieval corpus $\mathbb{D}$. Each instance involves employing natural language instructions to prompt the LLM to generate a paraphrase or description \citep{xrag}.

For Completion Pretraining, we randomly split documents from the RedPajama-Data-V2 \citep{redpajama} "2023-06" snapshot into two segments, where the length of the second segment is randomly sampled from the range [5, 100] to simulate realistic generation lengths.

\subsubsection{Details for Instruction Tuning Dataset}

We utilize the same instruction fine-tuning dataset as xRAG \citep{xrag}. Table \ref{summary_table} provides a summary, and Table \ref{detailed_table} offers detailed information about each subtask within the dataset. For question-answering tasks originally lacking explicit context, we employ Contriever \citep{contriever} to perform retrieval on the corpus $\mathbb{D}$, selecting the top-10 documents to serve as context.

\begin{table*}[htbp!]
\centering
\caption{Overall statistics of Instruction Tuning dataset.}
\begin{tabular}{@{}lclcc@{}}
\toprule
Task Type             & \# Involved datasets & \# Train & \# Prompt & \# Label  \\ \midrule
Reading Comprehension & 7                   & 488,344  & 447.62        & 30.34        \\
Summarization         & 3                   & 81,821   & 483.49        & 53.29        \\
Open Domain QA        & 7                   & 385,173  & 203.55        & 20.09        \\ \bottomrule
\end{tabular}
\label{summary_table}
\end{table*}

\begin{table*}[htbp!]
\centering
\caption{Detailed data statistics for our Context-aware Instruction Tuning Dataset.}
\begin{tabular}{@{}clccc@{}}
\toprule
\textbf{Task Type} &
  \multicolumn{1}{c}{\textbf{Dataset}} &
  \textbf{\# Train} &
  \textbf{\# Prompt Len} &
  \textbf{\# Label Len} \\ \midrule
\multirow{7}{*}{\begin{tabular}[c]{@{}c@{}}Reading \\ Comprehension\end{tabular}} &
  CoQA \citep{reddy2019coqa} &
  7101 &
  617.98 &
  77.75 \\
                     & DROP \citep{dua2019drop}       & 76098  & 356.06 & 3.86  \\
                     & NarrativeQA \citep{kočiský2017narrativeqa}  & 32747  & 702.39 & 7.86  \\
                     & PubMedQA  \citep{jin-etal-2019-pubmedqa}    & 1000   & 397.91 & 65.4  \\
                     & QuAIL \citep{rogers2020getting}         & 10246  & 512.9  & 2.0   \\
                     & SQuAD v2 \citep{rajpurkar2018know}         & 130319 & 214.54 & 6.87  \\
                     & PwC \citep{ge2023incontext}           & 241564 & 571.35 & 53.07 \\ \midrule
\multirow{7}{*}{\begin{tabular}[c]{@{}c@{}}Open Domain\\ QA\end{tabular}} &
  NQ \citep{natural} &
  87925 &
  203.62 &
  5.976 \\
                     & TriviaQA \citep{triviaqa}     & 78785  & 216.1  & 6.49  \\
                     & CommonsenseQA \citep{talmor2019commonsenseqa}& 9741   & 223.64 & 2.0   \\
                     & WikiQA \citep{yang-etal-2015-wikiqa}        & 1040   & 192.89 & 40.79 \\
                     & YahooQA      & 87358  & 196.56 & 56.7  \\
                     & FreebaseQA \citep{jiang-etal-2019-freebaseqa}    & 20353  & 218.49 & 4.87  \\
                     & MSMarco \citep{bajaj2018ms}       & 99994  & 194.82 & 15.91 \\ \midrule
\multicolumn{1}{l}{\multirow{3}{*}{Summarization}} &
  CNN/DM \citep{see2017point} &
  100000 &
  616.99 &
  63.37 \\
\multicolumn{1}{l}{} & SamSum \citep{Gliwa_2019}        & 14731  & 187.87 & 29.12 \\
\multicolumn{1}{l}{} & DialogSum \citep{chen2021dialogsum}   & 12460  & 247    & 37.61 \\ \bottomrule
\end{tabular}

\label{detailed_table}
\end{table*}

\subsubsection{Evaluation Dataset}

To ensure a comprehensive evaluation, we assess our method using the following 5 Open-Domain QA and 2 multihop QA:

\begin{itemize}
    \item \textbf{NaturalQuestions (NQ)} \citep{natural} contains questions corresponding to Google search queries. The open-domain version of this dataset is obtained by discarding answers with more than 5 tokens, each accompanied by a Wikipedia article containing the answer.
    \item \textbf{TriviaQA (TQA)} \citep{triviaqa} contains questions gathered from trivia and quiz-league websites. The unfiltered version of TriviaQA is used for open-domain question answering, each question is accompanied by pages from web and Wikipedia searches that may contain the answer.
    \item \textbf{WebQuestions (WQ)} \citep{wq} contains questions from web queries matched to corresponding entries in FreeBase.
    \item \textbf{PopQA (PQA)} \citep{popqa} focuses on factual question answering, posing challenges that test the
    model’s ability to recall precise knowledge and navigate ambiguities in entity representation.
    
    \item \textbf{ComplexWebQuestions (CWQ)} \citep{cwq} entails answering complex, multi-step questions sourced from the web, further challenging the model’s capacity to retrieve and reason over extensive web content.
    \item \textbf{2WikiMultihopQA (2WIKI)} \citep{2wiki} is designed to evaluate a model’s capability in multi-hop reasoning by synthesizing information from multiple Wikipedia passages.
\item \textbf{HotpotQA (HQA)} \citep{hotpotqa} similarly targets multi-hop reasoning, requiring models to amalgamate information from various contexts to answer a single query.

\end{itemize}

\subsection{Implementations}
\label{sec:imp}
Our implementations are based on Huggingface Transformers v4.45.2 \citep{transformers} using PyTorch v2.3.0 \citep{pytorch} and deepspeed\footnote{\url{https://github.com/microsoft/DeepSpeed}} v0.14.0. All experiments were conducted on 8 A100 NVIDIA GPUs, each equipped with 80GB of memory. In Table \ref{table:detail_pretrain} and Table \ref{table:detail_finetune}, we list the hyperparameters for Pretraining and Instruction Tuning.

\begin{table*}[ht!]
    \centering
    \begin{minipage}{0.49\textwidth}
        \centering
        \caption{Hyperparameters for Pretraining.}
        \resizebox{\linewidth}{!}{
        \begin{tabular}{cc}
            \toprule
            \textbf{Hyperparameter} & \textbf{Assignment} \\
            \midrule
            query tokens number  & 16 \\
            k\% & 10\% \\
            optimizer & AdamW \\
            learning rate & 1e-4 \\
            lr scheduler type & linear \\
            warmup ratio & 0.03 \\
            weight decay & 0.0 \\
            epochs & 1 \\
            flash attention & True \\
            batch size & 4 \\
            gradient accumulation steps & 4 \\
            num GPUs & 8 \\
            max sequence length & 2048 \\
            max train samples & 1,000,000 \\
            \bottomrule
        \end{tabular}
        }
        \label{table:detail_pretrain}
    \end{minipage}
    \hfill
    \begin{minipage}{0.49\textwidth}
        \centering
        \caption{Hyperparameters for Instruction Tuning.}
        \resizebox{\linewidth}{!}{
        \begin{tabular}{cc}
            \toprule
            \textbf{Hyperparameter} & \textbf{Assignment} \\
            \midrule
            query tokens number  & 16 \\
            k\% & 10\% \\
            optimizer & AdamW \\
            learning rate & 2e-5 \\
            lr scheduler type & linear \\
            warmup ratio & 0.03 \\
            weight decay & 0.0 \\
            epochs & 1 \\
            KL $\alpha$ & 2.0 \\
            KL temperature & 1.0 \\
            flash attention & True \\
            batch size & 4 \\
            gradient accumulation steps & 4 \\
            num GPUs & 8 \\
            max sequence length & 4096 \\
            max train samples & 955,338 \\
            \bottomrule
        \end{tabular}
        }
        
        \label{table:detail_finetune}
    \end{minipage}
\end{table*}

\subsection{Information Preservation Metrics}
\label{apmetric}

\textbf{BERTScore} is a metric used to evaluate the semantic similarity between a compressed text and its source. Unlike traditional metrics that rely on surface-level n-gram matching, BERTScore leverages contextual embeddings from models like BERT to compute similarity at the semantic level.

\textbf{Information Loss} quantifies the amount of information from the original text that is not successfully retained in the compressed text. A lower information loss indicates a more effective compression method in terms of preserving content.
Information quantity can be measured using the concept of Entropy ($H$). Higher entropy generally corresponds to higher information content. The information loss is defined as the difference between the information content of the original text $\boldsymbol{x}$ and the compressed text $\hat{\boldsymbol{x}}$, i.e., $H_{\boldsymbol{x}} - H_{\hat{\boldsymbol{x}}}$. 

\textbf{ROUGE} is a widely used set of metrics for evaluating the quality of automatically generated text summaries by comparing them to reference summaries (in this context, comparing the compressed text to the source or a gold standard summary derived from it). It primarily measures the overlap of units like n-grams or sequences between the compressed text and the original.

\textbf{Readability} assesses how easy a text is to read and understand. For compressed text, it measures the linguistic fluency and naturalness of the resulting output. Readability can be estimated using automated readability formulas, such as the Flesch Reading Ease score. These formulas typically consider factors like sentence length and the number of syllables per word to produce a score indicating reading difficulty.


\section{Limitations}
\label{limit}

While our proposed HyCo$_2$ demonstrates significant improvements in balancing local and global information retention for large language models (LLMs), several limitations warrant further investigation.

\noindent \textbf{Performance on Minimal Contexts (Top-1 Document)}: When processing only the top-1 retrieved document, particularly on certain IID datasets such as Natural Questions (NQ) and TriviaQA (TQA), HyCo2's performance may not consistently surpass xRAG. The hybrid architecture of HyCo2, designed to balance information from richer and more extensive contexts (e.g., top-3 or more documents), might be less optimized for these minimal input scenarios compared to approaches specifically tailored for single-document compression.

\noindent \textbf{Domain-Specific Generalization.} The current experiments primarily focus on knowledge-intensive question answering tasks, which limits the evaluation scope of HyCo$_2$ to specific domains. Future work should assess the framework's effectiveness across a broader range of applications such as code generation, legal document summarization, or technical report analysis, where context structure and relevance may differ significantly.

\noindent \textbf{Compression Granularity.} HyCo$_2$ retains approximately 10\% of input tokens by default through its classification layer, but this threshold is static and does not dynamically adapt based on content complexity or task-specific requirements. In some cases, particularly with highly nuanced or domain-specific texts, this fixed ratio might discard critical details essential for downstream reasoning.

\noindent \textbf{Latency in Long-Context Scenarios.} While HyCo$_2$ reduces token usage by an average of 88.8\%, the compression process itself introduces additional computational overhead due to the alternating training strategy and dual-path architecture. This can lead to increased latency during inference when dealing with extremely long contexts, potentially offsetting some efficiency gains.

\noindent \textbf{Scalability with Larger Models.} The current implementation has been tested on LLMs with parameter sizes up to 8B (e.g., LLaMA3.1-8B-Instruct). However, scaling HyCo$_2$ to handle ultra-large models (e.g., those exceeding 13B parameters) or multi-modal architectures could present new challenges in terms of memory footprint, adapter integration, and training convergence.

\noindent \textbf{Loss of Semantic Nuances.} Despite improved information preservation compared to existing methods like xRAG, soft compression via the hybrid adapter still risks losing subtle semantic nuances embedded in the original text. This limitation becomes more pronounced in contexts requiring deep inferencing, idiomatic understanding, or culturally specific interpretations.

\noindent \textbf{Dependency on Pretrained Components.} The effectiveness of HyCo$_2$ relies on the quality of underlying pretrained LLMs and their alignment with the hybrid adapter design. Performance may vary significantly when applied to less mature or low-resource language models, particularly for non-English or domain-specific architectures.

\section{Prompts Used in the Experiments}
\label{apprompt}

\subsection{Training}
\label{prompt_train}
To ensure consistency and clarity in pertraining and instruction tuning, we used several prompt templates as shown in Table~\ref{tab:paraphrase_instructions}, \ref{table:completion} and \ref{table:inst}.

\begin{table*}[ht!]\centering
\caption{Instructions used for Paraphrase Pretraining where \token and \doctoken are placeholders for projected feature $\mathcal{V}$ and document $\textrm{D}$ like \citep{xrag}.}
\begin{minipage}{0.99\columnwidth}\vspace{0mm}    \centering
\begin{tcolorbox} 
    \centering
     \hspace{-6mm}
\begin{itemize}[leftmargin=1mm]
\setlength{\itemsep}{2pt}
\item   "Background: \token means the same as \doctoken"
\item "Background: \token Can you put the above sentences in your own terms? \doctoken"
\item "\token Please provide a reinterpretation of the preceding background text. \doctoken"
\item "These two expressions are equivalent in essence:(1) \token (2) \doctoken"
\item "Background: \token is a paraphrase of what? \doctoken"
\item "\token Could you give me a different version of the background sentences above? \doctoken"
\item "In other words, background: \token is just another way of saying: \doctoken"
\item "You're getting across the same point whether you say background: \token or \doctoken"
\item "\token After unpacking the ideas in the background information above, we got: \doctoken"
\item "\token Please offer a restatement of the background sentences I've just read. \doctoken"
\item "Background: \token, which also means: \doctoken"
\item "Strip away the mystery, and you'll find \token is simply another rendition of: \doctoken"
\item "The essence of background: \token is captured again in the following statement: \doctoken"
\end{itemize}
\end{tcolorbox}
\vspace{-2mm}
    \label{tab:paraphrase_instructions}
\end{minipage}
\end{table*}

\begin{table*}[ht!]\centering
\caption{Instructions used for Completion Pretraining.}
\begin{minipage}{0.99\columnwidth}\vspace{0mm}    \centering
\begin{tcolorbox} 
    \centering
     \hspace{-6mm}
\begin{itemize}[leftmargin=1mm]
\setlength{\itemsep}{2pt}
 \item    "Using the background \token, generate a logical and coherent continuation paragraph.",
 \item    "Consider the background \token. Write the next paragraph that fits the context.",
 \item    "Based on the background  \token, draft a suitable continuation paragraph.",
  \item   "Referencing the background \token, create a seamless continuation.",
  \item   "Incorporate the background \token to generate the next segment of the text.",
  \item   "Leverage the background \token to produce the next logical section.",
  \item   "Using \token as the background, write the next paragraph.",
  \item   "Generate a follow-up paragraph that incorporates the background \token.",
  \item   "From the given background \token, create a continuation paragraph.",
  \item   "Background: \token",
 \item    "To provide accurate answers, it's essential to consider the background information presented here. Contextual Background: \token",
 \item    "Background: \token You might find the above background documents helpful.",
 \item    "The following background will help you understand the context for the questions. Please read it carefully before \item responding. Background: \token",
\end{itemize}
\end{tcolorbox}
\vspace{-2mm}
    \label{table:completion}
\end{minipage}
\end{table*}

\begin{table*}[ht!]\centering
\caption{Instructions used for Instruction Tuning.}
\begin{minipage}{0.99\columnwidth}\vspace{0mm}    \centering
\begin{tcolorbox} 
    \centering
     \hspace{-6mm}
\begin{itemize}[leftmargin=1mm]
\setlength{\itemsep}{2pt}
 \item    "Refer to the background document and answer the question. Provide only a short answer.
 Background: \token
 Question: \{question\}"
\end{itemize}
\end{tcolorbox}
\vspace{-2mm}
    \label{table:inst}
\end{minipage}
\end{table*}

\subsection{Reconstruct}
We use various prompting strategies to reconstruct original context from compressed representations \citep{improve}. As shown in Table~\ref{tab:reconstruction_prompts}, these prompts aim to encourage models to rephrase or expand latent semantic representations into natural language text. 

\begin{table*}[ht!]\centering
\caption{Prompts used for reconstructing contexts encoded by soft prompt compression method.}
\begin{minipage}{0.99\columnwidth}\vspace{0mm}    \centering
\begin{tcolorbox} 
    \centering
     \hspace{-6mm}
\begin{itemize}[leftmargin=1mm]
\setlength{\itemsep}{2pt}
\item   "These two expressions are equivalent in essence: (1) \token (2)"
\item "In other words, background: \token is just another way of saying:"
\item "Background: \token means the same as"
\item "\token After unpacking the ideas in the background information above, we got:"
\item "\token Please offer a restatement of the background sentences I've just read."
\end{itemize}
\end{tcolorbox}
\vspace{-2mm}
    \label{tab:reconstruction_prompts}
\end{minipage}
\end{table*}

\section{Ethical Considerations and AI writing statement}
 Our approach does not introduce ethical concerns. The datasets we used are public, and there are no privacy issues.

This paper utilized AI assistance for language polishing of the manuscript, including vocabulary correction and spell checking.


\newpage
\section*{NeurIPS Paper Checklist}

\begin{enumerate}

\item {\bf Claims}
    \item[] Question: Do the main claims made in the abstract and introduction accurately reflect the paper's contributions and scope?
    \item[] Answer: \answerYes{} 
    \item[] Justification: The abstract and introduction accurately reflect the paper's contributions and scope.

\item {\bf Limitations}
    \item[] Question: Does the paper discuss the limitations of the work performed by the authors?
    \item[] Answer: \answerYes{} 
    \item[] Justification: We can find the limitations in \ref{limit}.

\item {\bf Theory Assumptions and Proofs}
    \item[] Question: For each theoretical result, does the paper provide the full set of assumptions and a complete (and correct) proof?
    \item[] Answer: \answerNA{} 
    \item[] Justification:  Our paper does not include theoretical results.

    \item {\bf Experimental Result Reproducibility}
    \item[] Question: Does the paper fully disclose all the information needed to reproduce the main experimental results of the paper to the extent that it affects the main claims and/or conclusions of the paper (regardless of whether the code and data are provided or not)?
    \item[] Answer: \answerYes{} 
    \item[] Justification: We can reproduce the main experimental results following our settings in \ref{experiment} and \ref{main_ex}.

\item {\bf Open access to data and code}
    \item[] Question: Does the paper provide open access to the data and code, with sufficient instructions to faithfully reproduce the main experimental results, as described in supplemental material?
    \item[] Answer: \answerYes{} 
    \item[] Justification: We'll open source the code to an anonymous site \url{https://anonymous.4open.science/r/HyCo2} and put it on github after review.

\item {\bf Experimental Setting/Details}
    \item[] Question: Does the paper specify all the training and test details (e.g., data splits, hyperparameters, how they were chosen, type of optimizer, etc.) necessary to understand the results?
    \item[] Answer: \answerYes{} 
    \item[] Justification: We can find the experimental settings (hyperparameters and datasets) in \ref{base} and \ref{experiment}.

\item {\bf Experiment Statistical Significance}
    \item[] Question: Does the paper report error bars suitably and correctly defined or other appropriate information about the statistical significance of the experiments?
    \item[] Answer: \answerYes{} 
    \item[] Justification: We examined the effect of different hyperparameters on results in \ref{ab_s}.

\item {\bf Experiments Compute Resources}
    \item[] Question: For each experiment, does the paper provide sufficient information on the computer resources (type of compute workers, memory, time of execution) needed to reproduce the experiments?
    \item[] Answer: \answerYes{} 
    \item[] Justification: We can find it in \ref{analysis} and \ref{experiment}.
    
\item {\bf Code Of Ethics}
    \item[] Question: Does the research conducted in the paper conform, in every respect, with the NeurIPS Code of Ethics \url{https://neurips.cc/public/EthicsGuidelines}?
    \item[] Answer: \answerYes{} 
    \item[] Justification: All of our studies follow the NeurIPS Code of Ethics.

\item {\bf Broader Impacts}
    \item[] Question: Does the paper discuss both potential positive societal impacts and negative societal impacts of the work performed?
    \item[] Answer: \answerNA{} 
    \item[] Justification: There is no societal impact of the work performed.
    
\item {\bf Safeguards}
    \item[] Question: Does the paper describe safeguards that have been put in place for responsible release of data or models that have a high risk for misuse (e.g., pretrained language models, image generators, or scraped datasets)?
    \item[] Answer: \answerNA{} 
    \item[] Justification: The paper poses no such risks.

\item {\bf Licenses for existing assets}
    \item[] Question: Are the creators or original owners of assets (e.g., code, data, models), used in the paper, properly credited and are the license and terms of use explicitly mentioned and properly respected?
    \item[] Answer: \answerYes{} 
    \item[] Justification: We follow their open-source protocols in all our uses.

\item {\bf New Assets}
    \item[] Question: Are new assets introduced in the paper well documented and is the documentation provided alongside the assets?
    \item[] Answer: \answerNA{} 
    \item[] Justification: This paper does not release new assets.

\item {\bf Crowdsourcing and Research with Human Subjects}
    \item[] Question: For crowdsourcing experiments and research with human subjects, does the paper include the full text of instructions given to participants and screenshots, if applicable, as well as details about compensation (if any)? 
    \item[] Answer: \answerNA{} 
    \item[] Justification: This paper does not involve crowdsourcing nor research with human subjects.

\item {\bf Institutional Review Board (IRB) Approvals or Equivalent for Research with Human Subjects}
    \item[] Question: Does the paper describe potential risks incurred by study participants, whether such risks were disclosed to the subjects, and whether Institutional Review Board (IRB) approvals (or an equivalent approval/review based on the requirements of your country or institution) were obtained?
    \item[] Answer: \answerNA{} 
    \item[] Justification: This paper does not involve crowdsourcing nor research with human subjects.

\item {\bf Declaration of LLM usage}
    \item[] Question: Does the paper describe the usage of LLMs if it is an important, original, or non-standard component of the core methods in this research? Note that if the LLM is used only for writing, editing, or formatting purposes and does not impact the core methodology, scientific rigorousness, or originality of the research, declaration is not required.
    \item[] Answer: \answerNA{} 
    \item[] Justification: LLM is only used for writing and does not involve method implementation and innovation.

\end{enumerate}

\end{document}